\documentclass[icol]{sn-jnl} 

\usepackage{amssymb}
\usepackage{amsmath}
\usepackage{geometry}

\newcommand{\R}{\mathbb{R}}

\newcommand{\etal}{\textit{et al}. }

\newcommand{\Mesh}{\mathcal S}

\begin{document}

\title{Self Supervised Networks for Learning Latent Space Representations of Human Body Scans and Motions}

\author[1]{\fnm{Emmanuel} \sur{Hartman}}
\author[2]{\fnm{Martin} \sur{Bauer}}
\author[1]{\fnm{Nicolas} \sur{Charon}}

\affil[1]{\orgdiv{Department of Mathematics}, \orgname{University of Houston}, 
  \orgaddress{\street{641 Philip Guthrie Hoffman Hall, 3551 Cullen Blvd}, \city{Houston}, 
  \state{Texas}, \postcode{77021}, \country{USA}}}

\affil[2]{\orgdiv{Department of Mathematics}, \orgname{Florida State University}, 
  \orgaddress{\street{208 Love Building, 1017 Academic Way}, \city{Tallahassee}, 
  \state{Florida}, \postcode{32304}, \country{USA}}}

\abstract{%
This paper introduces self-supervised neural network models to tackle several fundamental problems in the field of 3D human body analysis and processing. First, we propose VariShaPE (Varifold Shape Parameter Estimator), a novel architecture for the retrieval of latent space representations of body shapes and poses. This network offers a fast and robust method to estimate the embedding of arbitrary unregistered meshes into the latent space. Second, we complement the estimation of latent codes with MoGeN (Motion Geometry Network) a framework that learns the geometry on the latent space itself. This is achieved by lifting the body pose parameter space into a higher dimensional Euclidean space in which body motion mini-sequences from a training set of 4D data can be approximated by simple linear interpolation. Using the SMPL latent space representation we illustrate how the combination of these network models, once trained, can be used to perform a variety of tasks with very limited computational cost. This includes operations such as motion interpolation, extrapolation and transfer as well as random shape and pose generation.
}


\maketitle

\section{Introduction}
The examination of human body shapes holds significant importance in various fields such as computer vision, graphics, and virtual reality.
This paper introduces new self-supervised deep learning models which, together, can be used to tackle some of the most common challenges in the field of human body shape analysis under minimal requirements on mesh preprocessing. Our main contributions are two-fold:
\begin{enumerate}
\item First, we introduce {\bf VariShaPE} (Varifold Shape Parameter Estimator), a self-supervised neural network model for real-time latent space encoding of unregistered human body scans; 
\item Second, we derive {\bf MoGeN} (Motion Geometry Network), a framework for learning the geometry of human body motion latent spaces from 4D-data. 
\end{enumerate}
Direct applications of the proposed algorithms include:
\begin{enumerate}
    \item A learning based framework for {\bf accurate motion transfer, motion interpolation and motion extrapolation}; 
    \item Fast algorithms for {\bf clustering and classification} of (large scale) datasets of unregistered human body scans;
    \item {\bf Generative modeling} via empirical distributions learned from (large scale) unregistered human body scan data bases. 
    \end{enumerate}
In our experimental section we demonstrate the capabilities of our framework on the DFAUST dataset and by relying on the the SMPL model (A Skinned Multi-Person Linear body representation)~\cite{SMPL:2015} as the corresponding latent space representation. We want to emphasize, that our framework is not tied in any way to this particular latent space model, but could be directly applied to any other existing latent space representation for human bodies, such as SMPL-X~\cite{SMPL-X:2019}, STAR~\cite{STAR:2020}, BLISS~\cite{muralikrishnan2023bliss} or the elastic shape analysis basis representation used in~\cite{Hartman_2023_ICCV,hartman2025basis,Pierson_2022_WACV}. We decided to focus the experiments on the SMPL representation, due to its convenient handling in PyTorch and its wide use in the the field of 3D human body shape community. In one of our ablation studies, cf. Section~\ref{sec:ablation}, we demonstrate a similar performance using the elastic shape analysis basis representation as an alternative latent space; in future work we plan to explore the use of some of the recent SMPL extensions, such as the above mentioned STAR~\cite{STAR:2020}. 
    
An additional benefit of our framework is the fast training time as compared to other network-based approaches in human shape analysis. This stems from our use of geometrically motivated constructions in the network, which allows us to encode fundamental invariances directly in the proposed architecture. Thereby we are able to significantly reduce the number of trainable parameters and thus require an order of magnitudes lower amount of training time as compared to previous approaches such as  LIMP~\cite{cosmo2020limp}, ARAPReg~\cite{huang2021arapreg}, or 3D-Coded~\cite{groueix20183d}.
\subsection*{Acknowledgements and Conflicts of Interest}
M.B was partially supported by NSF grant CISE 2426549. N.C and E.H were partially supported by NSF grants NSF CISE-2426550 and NSF DMS-2438562.

The authors certify that they have no conflicts of interest. 
\section{Related Work}
\subsection{Latent Space Models for Human Body Scan Representations}
Finding low-dimensional parametric representation for the statistical analysis of human shapes is an active research area~\cite{anguelov2005scape, hasler2009statistical, pishchulin2017building}, that dates back to the groundbreaking work on 3D Morphable Models for faces by Blanz~\etal~\cite{blanz1999morphable}. In particular applications to whole human body scans, common approaches include skeleton based apporaches in combination with physically motivated transformation models. Among those, one of the most widely used is the SMPL latent space~\cite{SMPL:2015}, which is learned from thousands of 3D body scans; see also~\cite{SMPL-X:2019,STAR:2020} for recent extensions that further enhance this model by adding e.g. articulated hands and face expressions.  

Our first contribution deals with the challenging task of extracting such latent space representations given raw data, i.e., to retrieve the best set of latent parameters to fit a given unregistered scan of a human body. Standard methods for this retrieval task are usually optimization based and often require manual intervention in order to lead to satisfactory results~\cite{SMPL:2015}; this includes spectral based methods such as the functional map framework~\cite{ovsjanikov2012functional}, minimizing the Hausdorff or Chamfer distance~\cite{fan2017point, groueix20183d}, or methods based on Riemannian geometry and elastic shape analysis~\cite{Hartman_2023_ICCV,hartman2025basis,JermynECCV2012,srivastava2016functional,hartman2023elastic}. In contrast to these methods, our model employs a self-supervised network approach making use of tools from geometric measure theory~\cite{kaltenmark2017general}, which allows us to incorporate shape invariances directly in the network architecture and thereby significantly boost the performance of the corresponding algorithms. 

\subsection{Interpolation in Latent Spaces}
Once data (e.g. human body scans) is embedded into a linear (Euclidean) latent space, one directly obtains a corresponding framework for statistical analysis: operations such as interpolation and extrapolation can be simply defined by performing these operations in the linear latent space. In the context of human body scans, this raises the issue of whether the linear geometry of the latent space accurately describes the complex movements and deformations that can be found, see also~\cite{michelis2021linear,mi2021revisiting,struski2023feature} for a discussion on this matter in the context of latent space interpolation for general machine learning applications.

Indeed, the performance of such a naive approach is rather limited when it comes to large movements of human bodies. To address these issues, multiple (physically motivated) deformation energy losses have been introduced in the training phase, thereby altering the corresponding latent space: this includes geodesic distances~\cite{cosmo2020limp}, the As-Rigid-As-Possible (ARAP) energy~\cite{huang2021arapreg, GLASSCvpr2022}, or volumetric constraints~\cite{atzmon2021augmenting}. See also~\cite{10.1007/978-3-031-20065-6_33, freifeld2012lie} for an approach using manifold regularization of learned pose spaces. These geometric constraints, however, significantly increase the total training costs of those approaches. More importantly, they are still based on user-defined assumption to model the  movement of human shapes. Our work differs in two different aspects from these approaches: first, we keep the latent space unchanged, but instead equip it with a different non-linear geometry. Secondly, we do not make any assumption on the physics behind the deformations of human body motions, but instead learn them in a purely data-driven approach using 4D training data. In our experiments, we argue that this data driven approach leads to a significantly improved performance in terms of interpolation and extrapolation accuracy. 

\section{Methods Section}
In the exploration of human body surfaces, a prevalent strategy for tackling the dimension reduction problem involves adopting a latent space model. This model is essentially a mapping $F$ from a latent space $\mathcal L\subset\R^n$ to the space of parameterized surfaces $\Mesh$ (triangulated meshes, respectively). Various construction methods, such as data-driven basis construction or skinned skeletal models, can be employed to build such latent spaces, see~\cite{SMPL:2015,SMPL-X:2019,STAR:2020,muralikrishnan2023bliss,Pierson_2022_WACV} and the references therein. The first focus of this paper is to solve the \textit{latent code retrieval problem,}, that is, for an arbitrary mesh $q$ and a given latent space model $F:\mathcal L\to \Mesh$ we aim to find $v\in\R^n$ such that $F(v)$ and $q$ represent the same shape. Whereas, the second contribution aims to employ a data-driven approach to learn the geometry of a given latent space representation based on real human body motions. 

Towards the first goal, the latent code retrieval problem, our approach is to train a fully connected neural network, $\Psi_{\theta}$, parameterized by weights $\theta=\{\theta_i\}$ mapping a mesh invariant feature vector representation of the input data to $\mathcal{L}$. This approach relies on an appropriate construction of mesh invariant representation of the input and comes with several notable challenges:
\begin{enumerate}
\item {\bf Shape preserving transformations:} the representation of a human body as a triangulated mesh comes with several ambiguities, i.e., many meshes can represent the same human scan and thus one has to compare these representations up to shape preserving transformations. These shape preserving transformations are given by translation, rotation and (most importantly) general remeshing operations such as subdivison or upsampling.   
\item {\bf Varying dimension and noise of real mesh data:} furthermore, meshes from real applications usually lack a fixed dimension. However, this is a prerequisite for the direct application of the proposed neural network architecture. In addition, these scans can exhibit a significant amount of imaging noise, such as holes or even larger missing parts. Our feature vector representation should remain robust to theses types of imaging noise.
\end{enumerate}
Before we are able to introduce our methods in more details, we will introduce a concept from geometric measure theory, which will be the basis for our mesh invariant feature vector construction.

\subsection{The Varifold Distance and its Gradient}
Many works in the field of human body analysis, including the aforementioned references~\cite{SMPL:2015,fan2017point, groueix20183d}, have relied on the well-known Chamfer or variants of the Hausdorff distance to define loss functions for surface reconstruction or estimation of latent representations. However, these typically result in loss functions that are not differentiable at every point. Alternatives, which arguably lead to better regularity and stability properties, can be obtained via representations of surfaces in measure spaces equipped with kernel metrics \cite{charon2020fidelity}, among which the varifold framework has recently seen several applications to the field \cite{amor2022resnet,pierson_eccv_2022,Hartman_2023_ICCV}. In the present work, we also leverage varifolds for two specific purposes: 1) to obtain a reliable notion of surface distance so as to evaluate and benchmark our proposed algorithms and, more importantly, 2) to obtain a mesh invariant feature representation for the VariShaPE architecture of the next section.

Let us briefly recap the primary features of the varifold framework for 3D surface meshes, referring to \cite{charon2013varifold,kaltenmark2017general,charon2020fidelity} for more detailed presentations. Assume that $q$ and $q'$ are two triangulated 3D surfaces not necessarily registered, i.e. their number of vertices and triangles may differ and no prior correspondences are known. Denoting $\{f_1,\ldots,f_m\}$ and $\{f'_1,\ldots,f_{m'}\}$ the respective set of non-degenerate triangular faces in $S$ and $S'$, the (squared) varifold distance is computed by examining all possible pairs of faces in the following way:
\begin{align}
   d_{\operatorname{Var}}(q,q')^2 =& \sum_{i=1}^{m} \sum_{j=1}^{m} \rho(|c_{f_i} - c_{f_j}|) \frac{(n_{f_i}\cdot n_{f_j})^2}{|n_{f_i}| |n_{f_j}|} +  \sum_{i=1}^{m'} \sum_{j=1}^{m'} \rho(|c_{f'_i} - c_{f'_j}|) \frac{(n_{f'_i}\cdot n_{f'_j})^2}{|n_{f'_i}| |n_{f'_j}|} \label{eq:dVar}
\\
   &\qquad- 2 \sum_{i=1}^{m} \sum_{j=1}^{m'} \rho(|c_{f_i} - c_{f'_j}|) \frac{(n_{f_i}\cdot n_{f'_j})^2}{|n_{f_i}| |n_{f'_j}|}\nonumber
\end{align}
where for a face $f$ (of $q$ or $q'$), $c_f$ denotes its barycenter and $n_f$ is the normal vector of length equal to the area of $f$. The function $\rho$ defines a kernel function which we take, in this work, to be a centered Gaussian which variance can be interpreted as the spatial scale of the metric. Note that many other choices of kernels on the face centers and normals are also possible, cf. \cite{kaltenmark2017general}. 
\begin{figure*}
\centering
    \includegraphics[width=.6\textwidth]{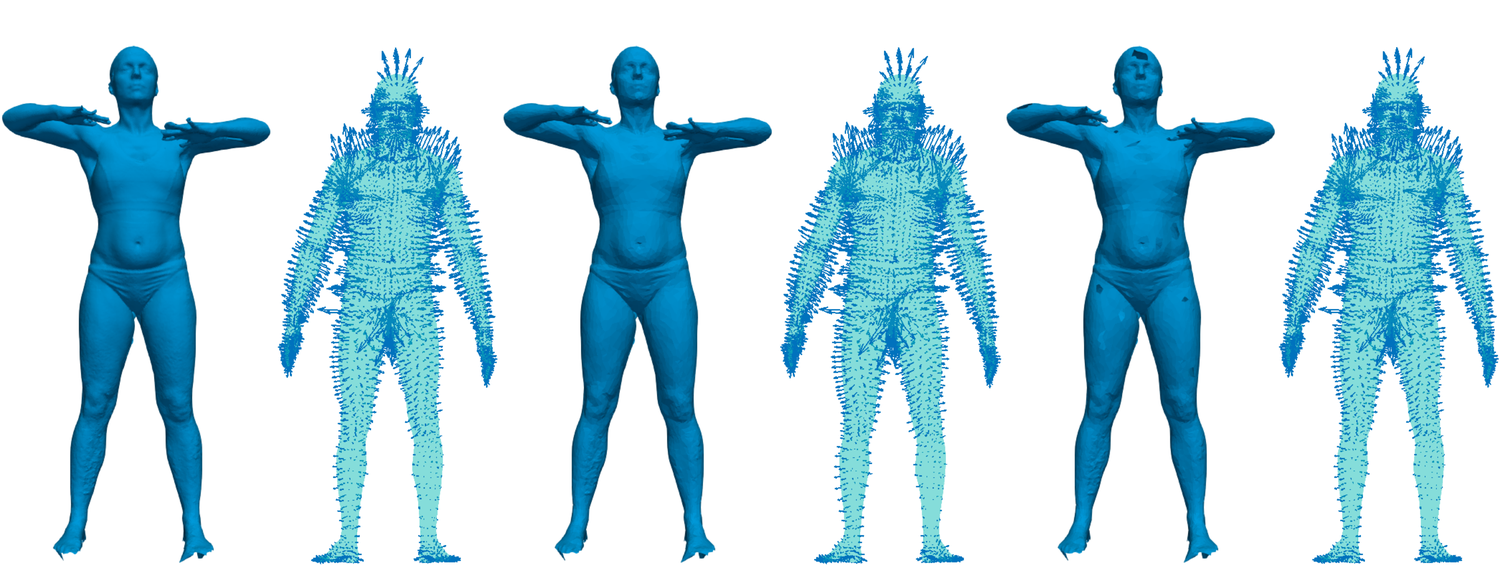}
    \caption{Plots of the varifold distance gradient field from the neutral pose template surface paired with three different target surfaces $q'$. These correspond to the same shape at full resolution (237k faces, left), lower resolution (20k faces, middle) and with added topological noise (right). One can see that the resulting field remains remarkably unaffected by those changes.} \label{fig:VarGrad}
\end{figure*}

We point out that the squared distance in \eqref{eq:dVar} is differentiable with respect to the vertex positions of $q$ and $q'$ provided the kernel function $\rho$ is smooth enough. The gradients can be easily computed either explicitly from \eqref{eq:dVar} or via automatic differentiation, with high numerical efficiency on a GPU due to the very parallelizable nature of this computation. Given a surface $q$, let us denote by $\nabla_{q'} d_{\operatorname{Var}}(q,\cdot)^2$ the distance gradient with respect to the second surface $q'$. The resulting vector field on $q'$, due to the properties of varifold metrics, is essentially invariant to resampling or mesh changes in the surface represented by $q$ as well as certain amount of imaging noise, which is showcased in the example of Fig. \ref{fig:VarGrad}. Thus it can be leveraged as an effective feature representation for latent space neural network learning models, as we explain in the next section.          

\subsection{Latent Code Retrieval: Mesh Invariant Feature Vectors and the Varifold Shape Parameter Estimation (VariShaPE) Framework }
\begin{figure*}
    \includegraphics[width=\textwidth]{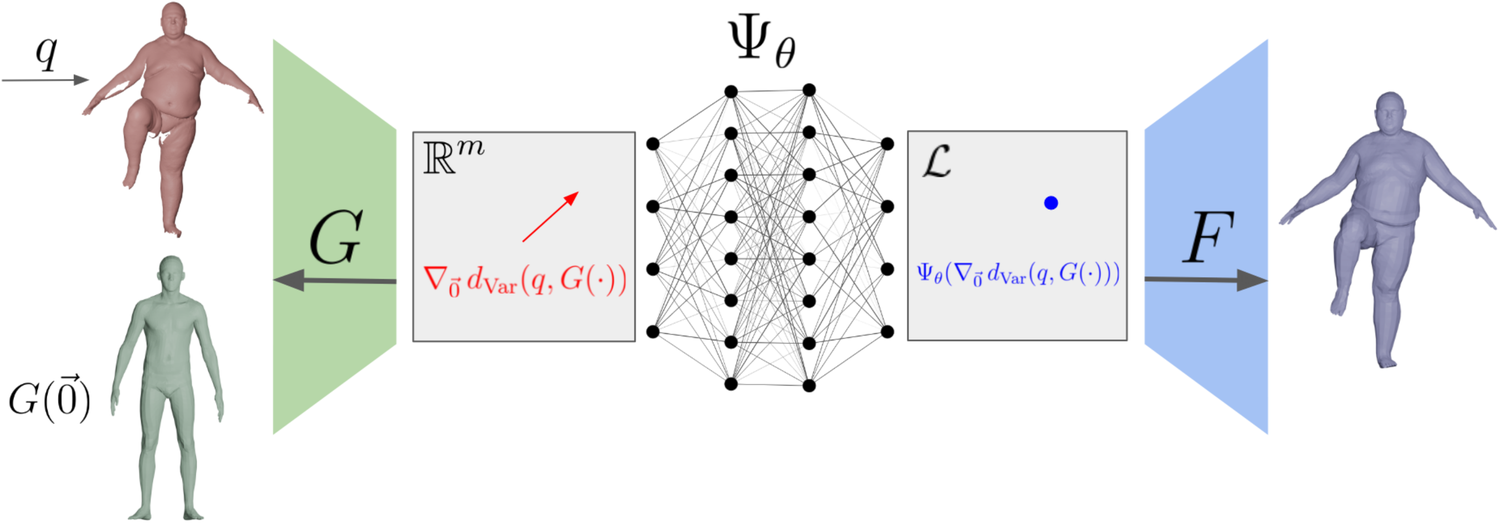}
    \caption{Illustration of the proposed VariShaPE framework: Given an input mesh $q$ with arbitrary mesh structure VariShaPE returns the estimated Latent Space Representantion. As can been seen on the input mesh our framework is robust to imaging noise, i.e., holes and small missing parts do not present any difficulties for the reconstruction procedure.} \label{fig:VariShaPE}
\end{figure*}
In recent years, several frameworks for sampling invariant feature vector representations of geometric data have been proposed. In the context of point clouds, PointNet~\cite{Charles_PointNet_2017} and Dynamic Graph Convolutional Neural Networks (DGCNN)~\cite{wang2019dynamic}, propose frameworks for constructing global features with fixed dimension representing this data. In the context of curves and shape graphs, the recent VariGrad model of~\cite{hartman2023VariGrad} utilizes a data driven method for feature vectors, which are obtained by fixing a template object and computing the gradient of the varifold distance with respect to the vertices of the template object following the setting described in the previous section. As compared to the PointNet and DGCNN architectures this approach led to a significantly improved performance for dealing with the resampling (remeshing, resp.) problem, which is the reason that we chose to adopt it in the present article.  

The drawback of this approach in the context of human body surfaces is that the template requires a large number of vertices to be sufficiently expressive. Consequently this produces a high dimensional feature vector representation and thus a higher computational complexity for training. To circumvent this difficulty we employ a second reduced dimension latent space model $G:\R^m\to \Mesh$ and instead compute the gradient with respect to the latent codes of $G$. It is important to note this additional latent model $G$ can be chosen independently of the targeted latent space representation $F$; it is merely a computational tool and (beyond computational accuracy) it is not influencing the final latent code retrieval. Indeed in our experiments, see Section~\ref{ssec:expeiremental_setup}, we choose different latent model for $F$ and $G$, cf. the ablation study of Section \ref{sec:ablation}, where we justify our particular choices.

The latent code retrieval algorithm we propose is thus the following: for an arbitrary mesh representing a human body, $q$, we take the varifold gradient at the template (w.r.t. to the latent codes of $G$) $\nabla_{\vec{0}} d_{\operatorname{Var}}(q,G(\cdot))^2$; this yield  a feature vector representation in $\R^m$, where $m$ is the dimension of the latent model $G$. Note that this representation inherits the low dimensionality from the model $G$ and the invariance to remeshing and robustness to noise from the varifold distance. These representations are then used as inputs to the fully connected network $\Psi_\theta$ that learns the latent code $v \in \mathcal L$ such that $F(v)$ and $q$ represent (approximately) the same shape. A schematic of this framework is presented in Figure~\ref{fig:VariShaPE}.

\begin{figure*}
    \centering
    \includegraphics[width=.9\textwidth]{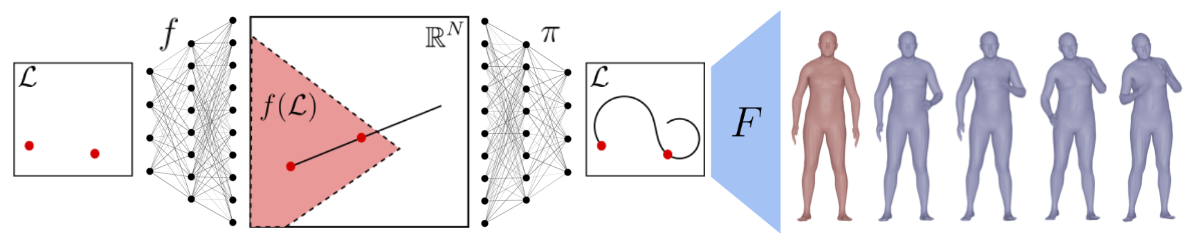}
    \caption{Illustration of the proposed MoGeN framework: Given two points in latent space (a point and a deformation direction, resp.), the network $f$ maps them into a high dimensional, lifted latent space, where interpolation (extrapolation, resp.) follow the standard linear paths. Subsequently, they are projected back to the original latent space using the network $\pi$, which allows the user to map them into the space of surfaces using the decoder $F$.} \label{fig:MoGeN}
\end{figure*}
\subsection{Learning the Latent Space Geometry}
In this section, we assume that our latent space representation can be decomposed into a Pose Parameter Space (i.e. those parameters that govern the motion/pose of the object) and a Shape Parameter Space (i.e. those parameters that govern the identity/body type of the object). This is a typical feature in the construction of human body latent spaces as is the case, for instance, with the SMPL latent space. As there is no ground truth for the deformation of one human body into another human body available we will equip the Shape Parameter Space simply with the linear Euclidean geometry.  The goal of this section is to learn the geometry of the Pose Parameter Space, so that human motions from 4D data are ``geodesics'' with respect to this geometry. Namely, our goal is to learn a ``geometry'' such that the following two operations correspond to natural human motions:
\begin{enumerate}
\item {\bf Interpolation} (the ``geodesic'' boundary value problem), which corresponds to finding the path (deformation) between two different latent codes .
\item {\bf Extrapolation} (the ``geodesic'' initial value problem), which corresponds to finding the trajectory given one latent code and an initial deformation direction. 
\end{enumerate}
To achieve this goal, we train a model comprised of two maps $f:\mathcal{L}\to \R^N$ and $\pi:\R^N\to \mathcal{L}$. 
Given a 4D sequence $\alpha_1,...,\alpha_T$ of latent codes representing a human motion, we define the loss function 
\begin{align*}
    &L(\alpha_1,...,\alpha_T)=\frac{1}{2T}\sum_{i=1}^T  \underbrace{\operatorname{MSE}(\hat{\alpha}_i,\alpha_i)}_{\text{Interpolation Loss}}+\underbrace{\operatorname{MSE}(\alpha_i',\alpha_i)}_{\text{Extrapolation Loss}}
\end{align*}
where $\operatorname{MSE}$ denotes the mean squared error and where
\begin{equation*}
    \hat{\alpha}_i=\pi\left(f(\alpha_0)+i\frac{f(\alpha_T)-f(\alpha_0)}{T-1}\right)
\end{equation*}
and
\begin{equation*}
  \alpha_i'= \pi(f(\alpha_0)+i(f(\alpha_1)-f(\alpha_0))).
\end{equation*}
Here, the first term is called \emph{Interpolation Loss} as it measures the deviation of the linear interpolation based on the first and last time point to the given data (i.e., the given human body mini sequence). On the other hand, the second term is called the \emph{Extrapolation Loss} as it measures the deviation of the linear extrapolation based on the first two time points of the given latent code sequence. It is important to note that our training ensures $\pi\circ f\approx\operatorname{Id}_{\mathcal{L}}$, but that we will not have $f\circ\pi=\operatorname{Id}_{\R^N}$ in general. The extrapolation loss term ensures in addition that $\pi$ learns to map elements of $\R^N$ which are not in $f(\mathcal{L})$ consistently with neighboring points in $f(\mathcal{L})$. This ensures that our framework learns appropriate behavior near the boundary of feasible human motions. We will also demonstrate the advantage of this feature in the experimental section, cf. Figure~\ref{fig:interp_extrap}.  For a schematic description of the general latent space geometry procedure, we refer to Fig.~\ref{fig:MoGeN}.



\section{Experimental Results}\label{sec:HBResults}
\subsubsection*{Experimental Setup:}\label{ssec:expeiremental_setup}
All experiments were performed on a standard home PC with a Intel 3.2 GHz CPU and a GeForce GTX 2070 1620 MHz GPU, where we used the following setups for our network architectures: for the VariShaPE model we choose $G$  as the affine decoder described in \cite{Hartman_2023_ICCV,hartman2025basis} and $F$ to be the base, gender-neutral SMPL model~\cite{SMPL:2015}. In Section \ref{sec:ablation} below, we perform an ablation study for different choices of latent space models for both $F$ and $G$. Consequently, $m=170$ and $n=375$. The network $\Psi_\theta$ is then constructed as sequence of fully connected layers separated by ReLU activation functions with a total of $3\times10^6$ trainable parameters.

Furthermore, we report results of the proposed MoGeN model for the latent space representation of motions with respect to the base, gender-neutral SMPL model. As mentioned above, the dimension of this latent space model is $375$. For the dimension of the lifted latent space we take $N=1500$. The maps $f,\pi$ are constructed as fully connected networks using ReLU activation functions and a combined $2\times10^7$ trainable parameters.

\begin{table*}[h]
    \centering
    \scriptsize
    \setlength\tabcolsep{2pt}     
    \renewcommand{\arraystretch}{1.0}
    \begin{tabular}{l|ccc|ccc|}
    &\multicolumn{3}{c|}{Registered}&\multicolumn{3}{c}{Unregistered}\\
    &Cham.&VAE&Ours&Cham.&VAE&Ours\\\hline
    Mean Vertex Dist.&0.063&0.051&\textbf{0.021}&NA&NA&NA\\
    Varifold Error&0.047&0.038&\textbf{0.035}&0.048&NA&\textbf{0.036}\\
    Chamfer Error&0.031&0.029&\textbf{0.025}&0.057&NA&\textbf{0.029}\\ \hline
    Time of $10^3$ Recon. &6339.4s&2.1s&0.8s&9311.6s&NA&1.0s\\
    Training&None&2w&8hrs&None&NA&8hrs\\
    \end{tabular}
    \qquad
    \begin{minipage}[h]{.30\linewidth}
    \centering
    \includegraphics[width=\textwidth]{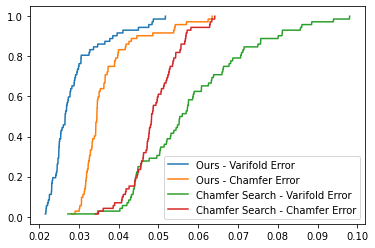} 
    \end{minipage}
    \caption{Reconstruction Results: We compute the mean vertex distance for the unseen registered testing data and mean Chamfer and varifold errors for all of the meshes in the registered and unregistered testing sets. For the unregistered data we display, on the right panel, the cumulative error distributions for each method and evaluation metric.\label{tab:registration}}
\end{table*}
\begin{figure*}
    \centering
    \begin{minipage}{.24\textwidth}    
    \vspace{-175pt}
    \renewcommand{\arraystretch}{5}
    \begin{tabular}{r}
    Chamfer Search:\\
    Input Data:\\
    Ours:
    \end{tabular}
    \end{minipage}
    \includegraphics[width=.75\textwidth]{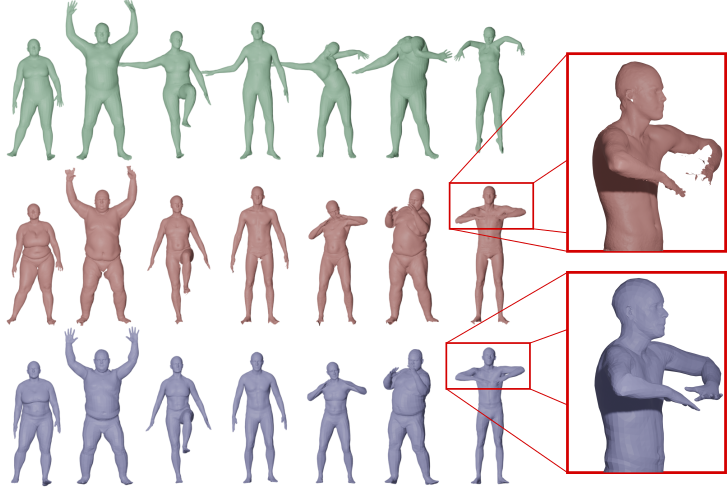}
    \caption{Qualitative Comparison of SMPL latent code retrieval methods for unregistered data: We display 7 raw scans of human meshes (middle), reconstructions produced by minimizing the Chamfer distance (top), and reconstructions produced via our model (bottom). Additionally, we highlight our model's robustness to noise and scanning errors, as can be seen in the highlighted box at the very right.} \label{fig:LatVG}
\end{figure*}
\subsubsection*{Datasets:} We train our networks on the Dynamic FAUST (DFAUST) dataset~\cite{dfaust:CVPR:2017}. This dataset contains high quality 4D scans captured at 60 Hz of 10 individuals performing 14 in-place motions. Due to the high speed of the recording, these scans have topological noise and scanning errors. These 4D sequences are registered using image texture information and a body motion model creating sequences of meshes with fixed point to point correspondences. Furthermore the meshes in these registered sequences share point to point correspondences with outputs of many latent space models such as \cite{SMPL:2015,SMPL-X:2019,Pierson_2022_WACV}. We then split this data so that 80 percent of the sequences are used in training and 20 percent are used in testing.  Additionally, for the selected testing sequences we also utilize the corresponding raw scans to evaluate the performance of the proposed \mbox{VariShaPE} model on meshes with topological noise and scanning errors and without fixed point to point correspondence. To train and validate the MoGeN model, we use the VariShaPE architecture to extract latent code representations of the training and testing sequences and subsequently extract mini-sequences of pose parameters from these sequences.

\begin{figure*}
    \centering
    \includegraphics[width=.55\textwidth]{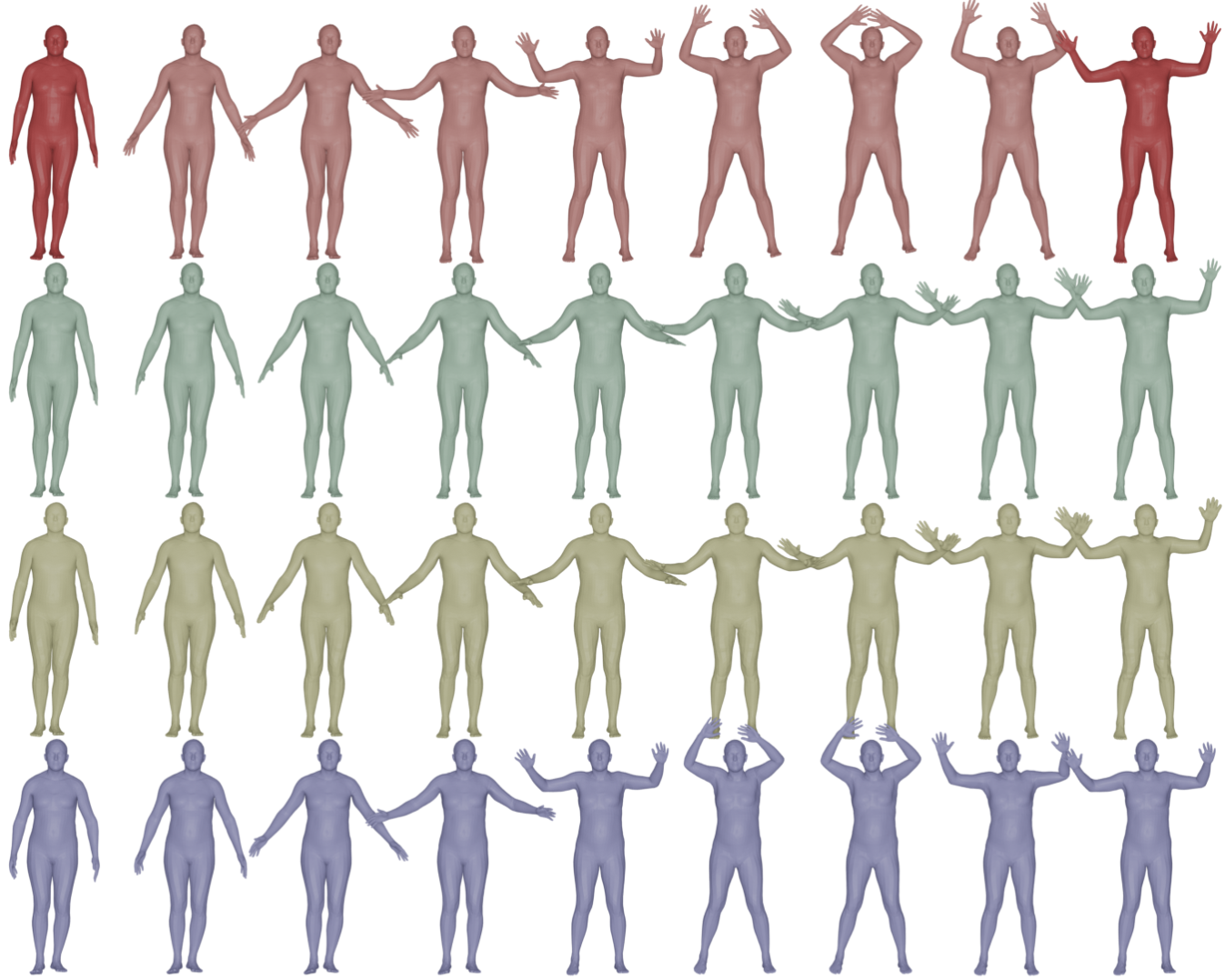}
    \qquad
    \includegraphics[width=.35\textwidth]{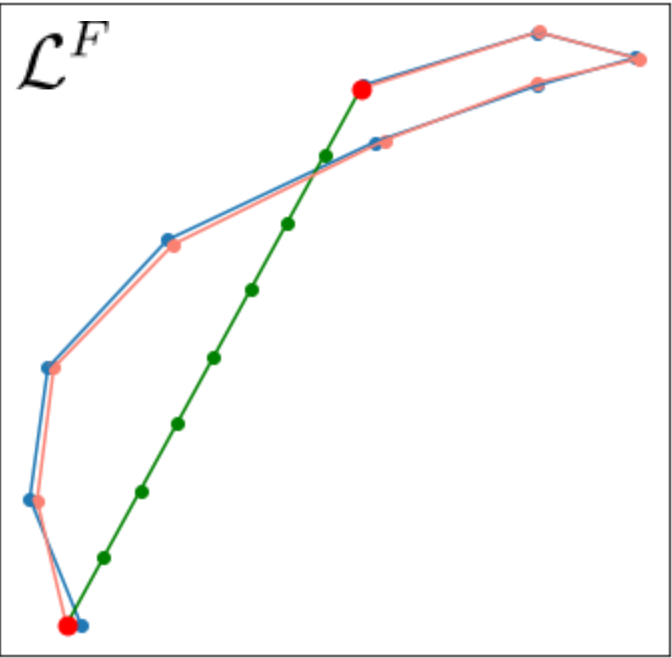}
    \caption{Qualitative comparison of MoGeN with linear interpolations in SMPL space. In red, we display a sequence of meshes corresponding to a human motion from data with the boundary points highlighted. In green, we display the path which arises from linear interpolation between the boundary points in the latent space $\mathcal L$. Finally, in blue, we display the path produced by the MoGeN interpolation model with the boundary points as inputs. On the right we visualize these paths in two coordinates of the latent space with the same color coding.} \label{fig:interp_extrap}
\end{figure*}
\begin{figure*}
    \centering
    \includegraphics[width=.95\textwidth]{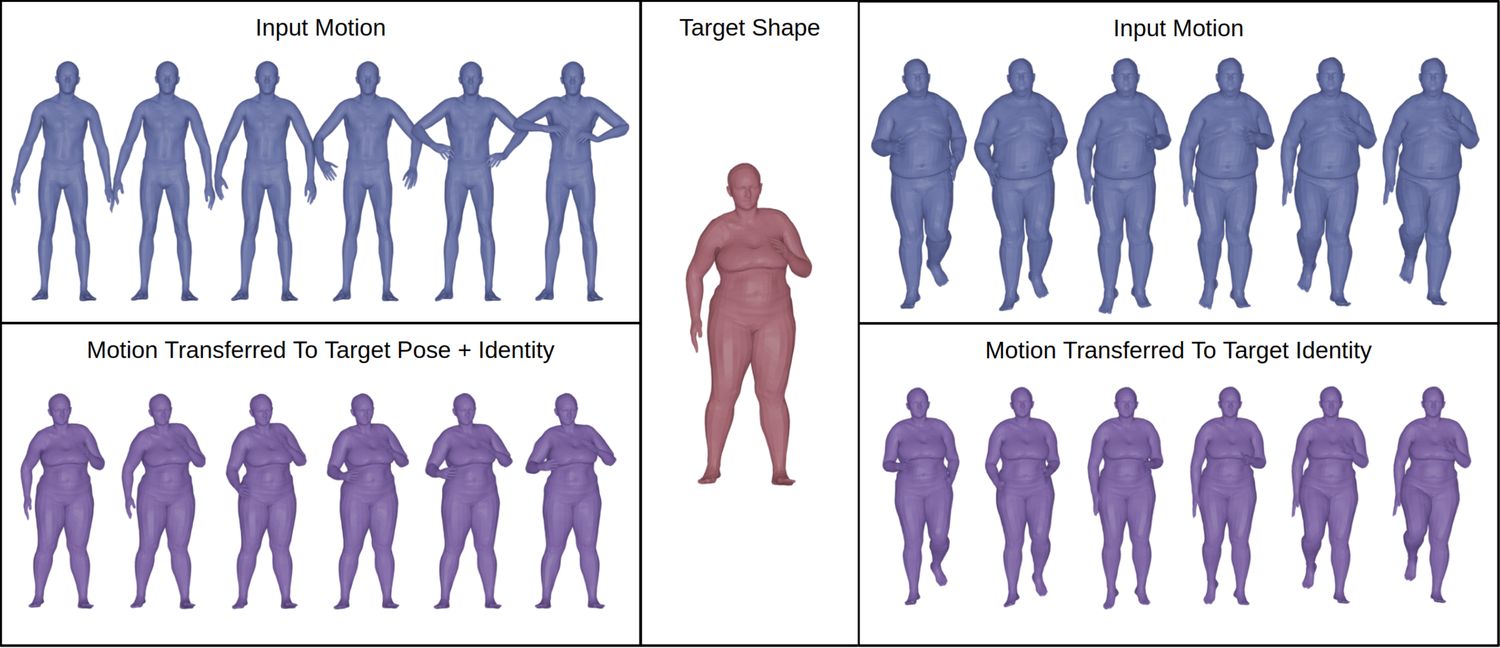}
    \caption{Examples of Motion Transfer: On the right we display the transfer of a motion from data (blue) to the identity of a target shape (red). Using the VariShaPE model we obtain latent codes for the target motion and shape. We then combine the pose parameters of the motion with the identity parameters of the target to produce an transferred motion (purple).  On the left we display the transfer of human motions from data (blue) onto the \textbf{\textit{pose and identity}} of the target shape (red).  Using MoGeN, we obtain lifted pose parameters for the sequence and target shape and apply a simple translation in the lifted space. We map the resulting path back to to the latent space and combine the transferred pose parameters with the target identity parameters producing the transferred motion (purple). } \label{fig:MT2}
\end{figure*}
 \begin{table*}
    \centering
    \scriptsize
    \setlength\tabcolsep{2pt}     
    \renewcommand{\arraystretch}{1.0}
    \begin{tabular}{l|ccc|ccc|}
    &\multicolumn{3}{c|}{Interpolation}&\multicolumn{3}{c}{Extrapolation}\\
   &Linear &\cite{huang2021arapreg}-AD&MoGeN&Linear&\cite{huang2021arapreg}-AD&MoGeN\\ \hline
    L$^2$ Vertex Error&0.0126&0.0103&\textbf{0.0046}&0.0142&0.0127&\textbf{0.0059}\\
    Varifold&0.0073&0.0069&\textbf{0.0060}&0.0092&0.0088&\textbf{0.0071}\\\hline
    Timing For $10^3$ Sequences&1.6s&2.2s&1.9s&1.6s&2.2s&1.9s\\
    Training Time&None&2w&12hrs&None&2w&12hrs\\
    \end{tabular}
    \caption{Interpolation and extrapolation results for SMPL model: We compute the mean vertex distance and mean varifold error for the testing set of mini-sequences extracted from the DFAUST testing set.}\label{tab:interp_extrap}
\end{table*}
 
\begin{figure*}
    \centering
    \includegraphics[width=\textwidth]{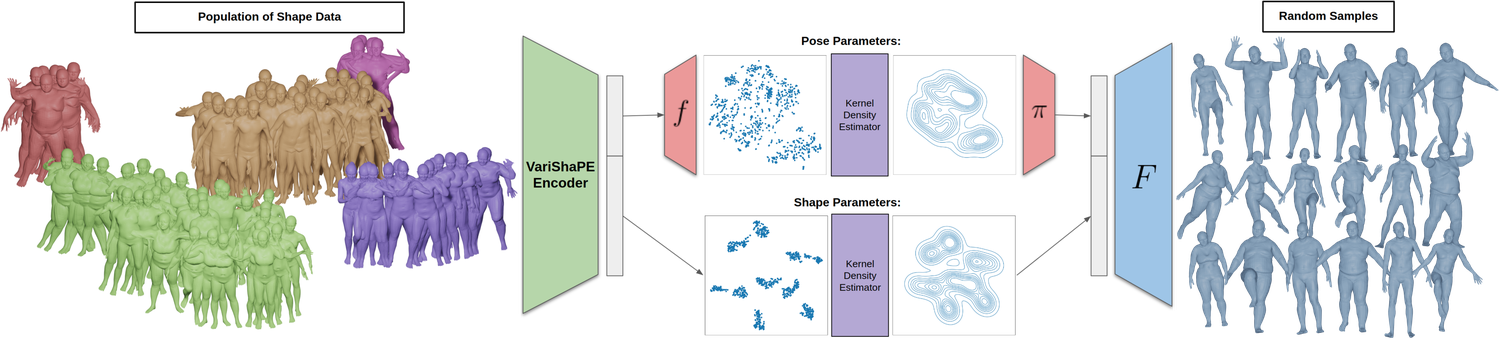}
    \caption{Here we describe the pipeline for a generative model based on our networks based on an empirical distribution of human bodies. First, we extract the pose and identity parameters of a dataset of human bodies using the VariShaPE model. We apply MoGeN to the distribution of pose parameters and fit a  kernel density estimator (KDE) in the lifted space. To generate a new pose we sample a code from this KDE and map it back to the latent space. We combine this with an identity code sampled from a KDE fit to the distribution of identity parameters to generate a random shape. } \label{fig:random}
\end{figure*}

\subsection{Mapping Registered and Unregistered Scans to SMPL}
In our first set of experiments we aim to demonstrate the capability of our \mbox{VariShaPE} architecture to learn a map from unregistered meshes into the latent space $\mathcal{L}$ of the SMPL model. As described in the setup section we train the model with  registered meshes from the DFAUST dataset. As such, we train the model to minimize the mean square error between the vertices of the training meshes and the reconstruction of the learned parameters of the model.
For comparison of our latent code retrieval capabilities we use two other state-of-the-art methods: first we run an optimization~\cite{fan2017point, groueix20183d} over the latent space using an L-BFGS optimizer minimizing the Chamfer distance between the target mesh and the reconstruction of the latent code. This method works for both registered and unregistered meshes as the Chamfer distance does not rely on point to point correspondences. Second, we compare the resulting reconstructions to the variational auto-encoder network of \cite{huang2021arapreg}. This approach is trained with registered data to construct a latent space representation of human bodies which is regularized by an as-rigid-as-possible (ARAP) energy. 
As this network is constructed for data with fixed point to point correspondences we cannot compare its performance on the unregistered testing set, but only use it for pre-registered data. A quantitative comparison for the latent code  retrieval task is presented in Tab.~\ref{tab:registration}, where we use three different measures to quantify the quality of a reconstruction: (1) the mean vertex distance (only available for registered meshes); (2) the Chamfer distance and (3) the varifold distance. As one can see VariShaPE significantly outperforms the competing methods, while at the same time having an order of magnitudes lower computational cost then the optimization based methods and a somewhat lower (but comparable) computational cost than the VAE based approach.  In addition to these quantitative results, we present qualitative results in Fig.~\ref{fig:LatVG}; here we restrict the comparison to Chamfer search as we focus the presentation to the more challenging task of retrieving the latent code directly from unregistered and noisy scans. See also Figure \ref{fig:ARAP} in Appendix A for a comparison to VAE.  
\begin{figure*}[hpt]
    \centering
    \includegraphics[width=.45\textwidth]{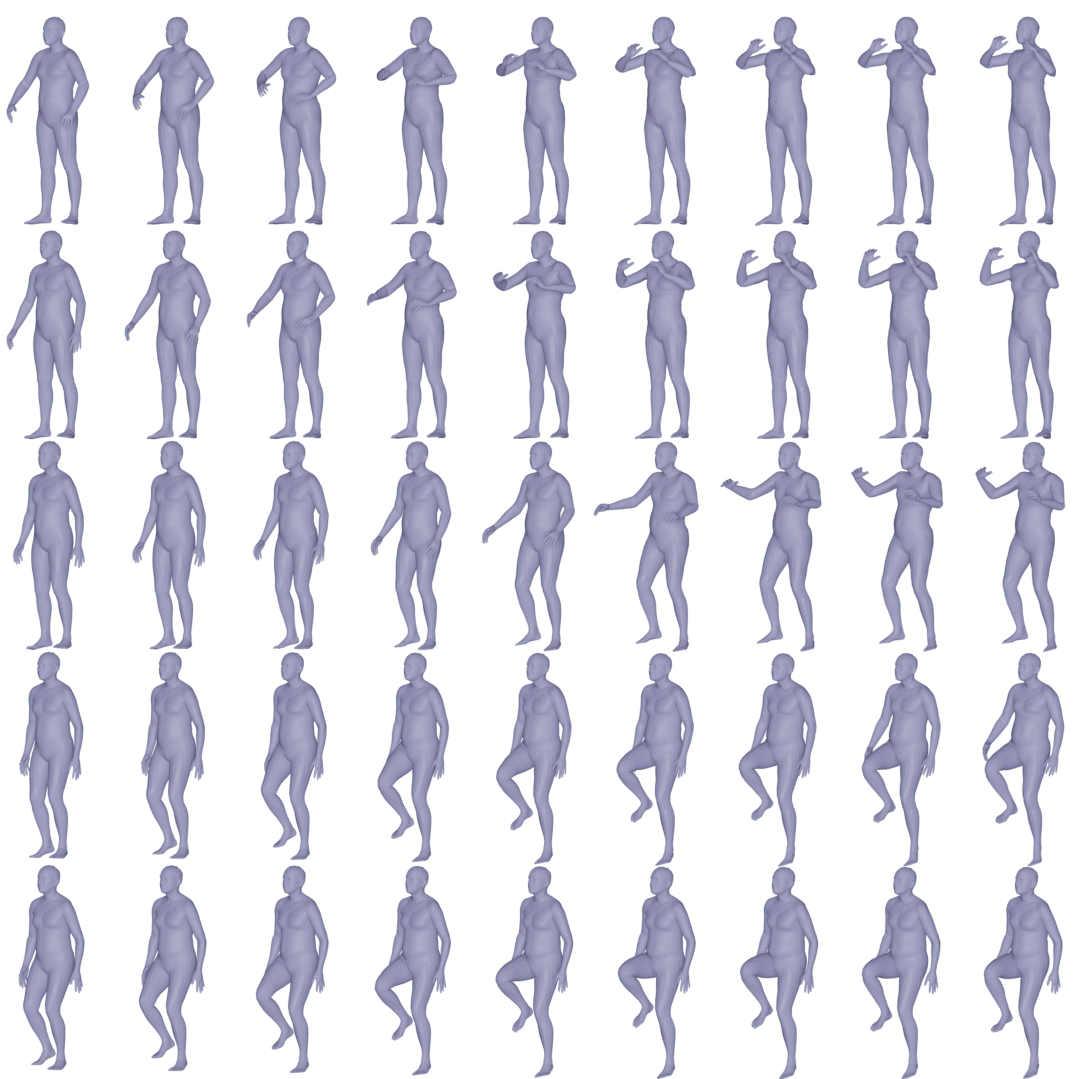}
    \includegraphics[width=.45\textwidth]{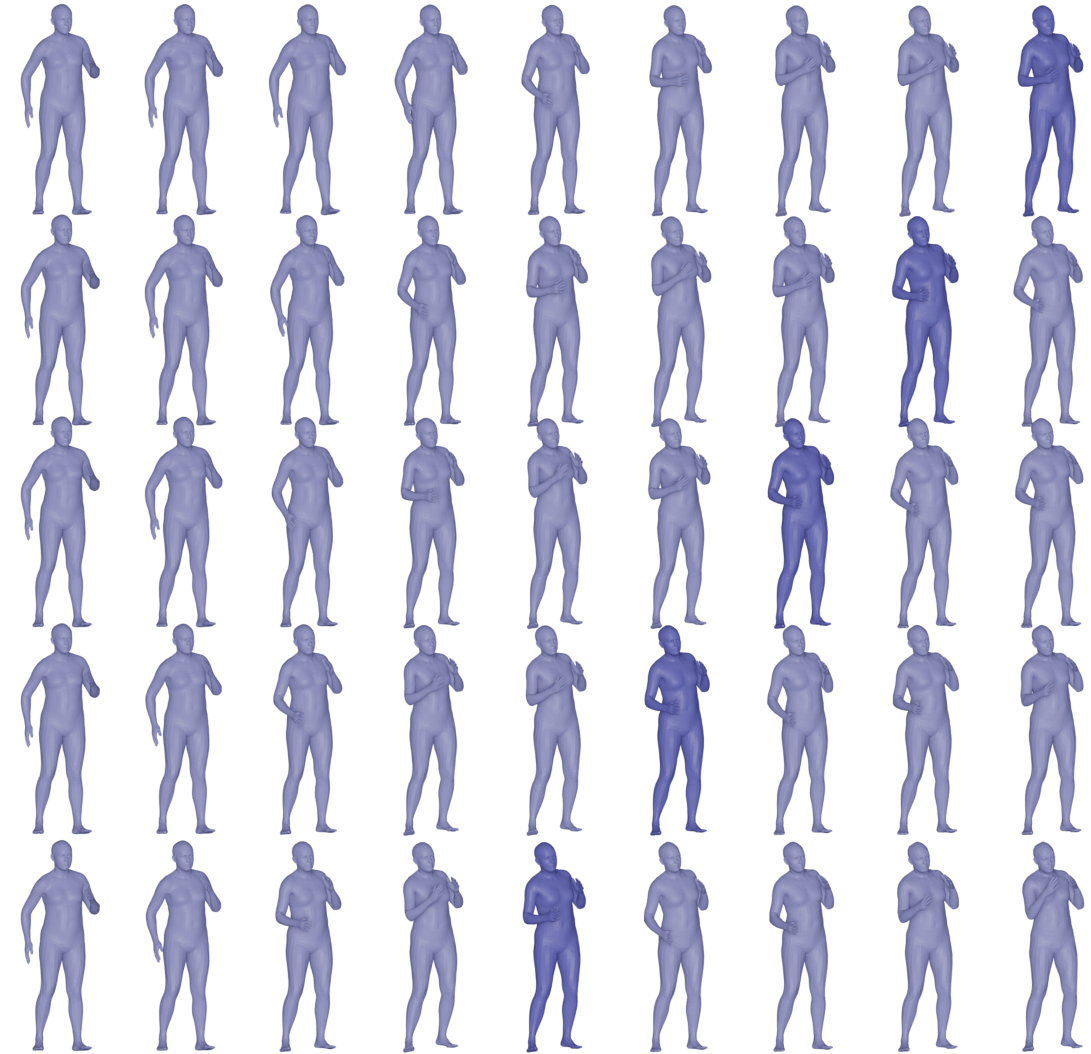}
    \caption{Interpolations between 4D data: We display two examples interpolations between 4D data using our models. The top and bottom rows correspond to motions from 4D data with the middle rows corresponding to the interpolation using our method. The example on the right demonstrates the capability of the algorithm to interpolate between motions, that are performed with a different speed: we show  the interpolation between the first half of a sequence to the full sequence, i.e., the second sequence performs the motion in half the time. Instead of blending the motion at each fixed time our algorithm correctly extends the length of the motion in the interpolation. The highlighted shape shows the moving endpoint of the first sequence.} \label{fig:4D}
\end{figure*}

\subsection{Motion Interpolation and Extrapolation:}
Next we investigate the capabilities of our second proposed network, MoGeN, to accurately learn the geometry of a latent space representations of human body motions. Therefore, we compare our method's ability to (1) predict human motions interpolating between a start and end point and (2) to extrapolate upon initial motions. To evaluate the performance of MoGeN we compare our results again to two competing methods: first, we compare to linear interpolations in the pose parameter space of the SMPL model. This is a natural baseline to compare our model against and allows us to visualize comparisons of our model in the latent space, cf. Figure~\ref{fig:interp_extrap} and Figure \ref{fig:extrap} in Appendix B. Secondly, we provide quantitative comparisons of the interpolations and extrapolations produced by our method and the auto-decoder latent space construction of \cite{huang2021arapreg}. In Tab.~\ref{tab:interp_extrap} one can observe that the proposed algorithm outperforms the competing methods, which is also visualized in the qualitative example in Figure ~\ref{fig:interp_extrap} and Figure \ref{fig:extrap}.

\subsection{Motion Transfer, Generative Modeling and Interpolation between 4D-data:} Having demonstrated the accuracy of the proposed network architectures, we aim to present three different applications of the proposed methodology. The first one consists of a real-time approach for motion transfer. Fig.~\ref{fig:MT2} shows two different examples and an explanation of the procedure that uses VariShaPE to encode the given sequence and target in latent space, and subsequently employs MoGeN to accurately transport the corresponding path of latent codes to the latent codes of the target shape.

Next, we demonstrate how one can combine MoGeN and VariShaPE to obtain a generative model for human bodies based on estimating an empirical distribution from observed training data, cf. Fig.~\ref{fig:random} for more details on the procedure. In that same figure, we show 18 random samples, where the empirical distribution was constructed using 30,000 scans. Note that this procedure is only feasible due to the light speed encoding capabilities for raw scans provided by VariShaPE: for our experiment the encoding of the whole population of unregistered training scans took less than one minute.

Finally, we show how our frameworks can be combined to lead to an algorithm for 4D-data interpolation (and extrapolation), i.e., we aim to interpolate (extrapolate, resp.) between two motion sequences $q_1(t)$ and $q_2(t)$. A simple algorithm for achieving this goal would consist of interpolating for each fixed time $t_0$ in the motion sequences between their corresponding latent codes. This can lead, however, to suboptimal results as humans perform motions at varying speed and thus $q_0(t_0)$ might not correspond to $q_1(t_0)$ but rather to a different time point $q_1(t_1)$. To address this challenge we resort to MoGeN, which maps motion sequences to straight lines in the lifted latent space, which reduces the 4D-interpolation problem to the task of averaging lines in $\mathbb R^N$. We present two examples of this procedure in Figure~\ref{fig:4D}, where the time independence of the 4D-interpolation algorithm is  highlighted in the example on the right.  
\begin{table*}
    \centering
    \scriptsize
    \setlength\tabcolsep{2pt}     
    \renewcommand{\arraystretch}{1.0}
    \begin{tabular}{l|cc|cc|}
    &\multicolumn{2}{c|}{Registered}&\multicolumn{2}{c}{Unregisterd}\\
    &Cham.&Ours&Cham.&Ours\\ \hline
    Mean Error&0.083&0.021&NA&NA\\
    Varifold Error&0.054&0.034&0.047&0.037\\
    Chamfer Error&0.026&0.025&0.032&0.030\\
    \end{tabular}\qquad
    \begin{tabular}{l|cc|cc|}
    &\multicolumn{2}{c|}{Registered}&\multicolumn{2}{c}{Unregisterd}\\
    &Cham.&Ours&Cham.&Ours\\ \hline
    Mean Error&0.063&0.022&NA&NA\\
    Varifold Error&0.047&0.037&0.048&0.041\\
    Chamfer Error&0.031&0.028&0.057&0.034\\
    \end{tabular}
    \caption{Ablation Study for Choices of Latent Space Models in the VariShaPE Model: In the left table, we present results for the VariShaPE model, with $F$ and $G$ chosen to be the latent space model of \cite{Hartman_2023_ICCV,hartman2025basis}. The right table contains the results for the VariShaPE model, with $F$ and $G$ chosen to be the latent space model of SMPL~\cite{SMPL:2015}. \label{tab:Ablation}}
\end{table*}
\begin{figure*}
    \centering
    \includegraphics[width=.95\textwidth]{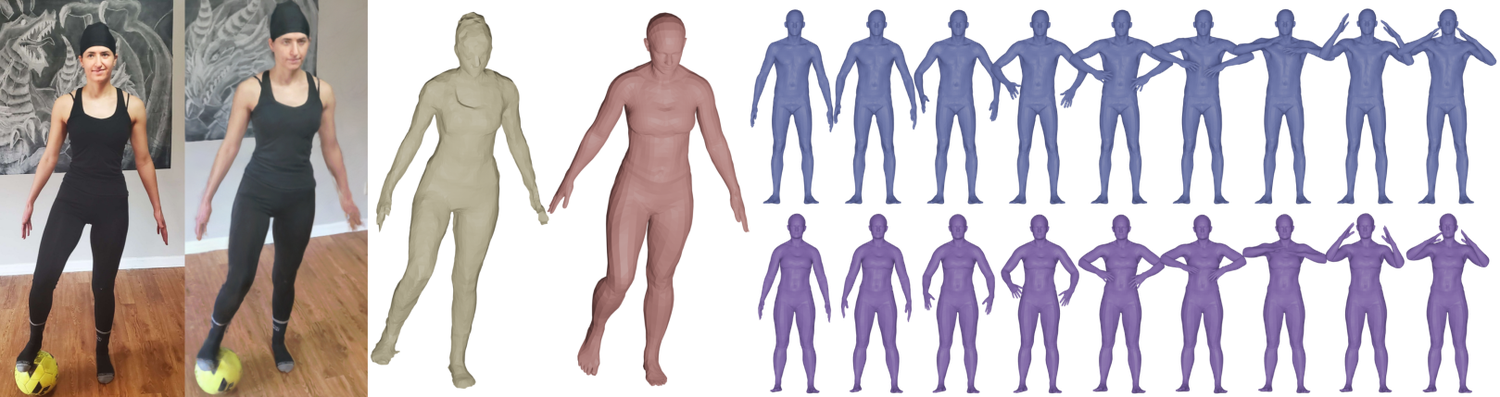}
    \caption{An example using Neural Radiance Field reconstruction from a body shape video scanned: On the left we show the retrieval of the latent code representation of a surface reconstructed from a Neural Radiance Field representation of the subject. On the right, we display a motion sequence on the top row and the transferred motion sequence with the selected identity.    } \label{fig:MT}
\end{figure*}

\subsection{Ablation Study using a different Latent Space Model}\label{sec:ablation}
In the previous part we reported the results of the VariShape model with $F$ as the SMPL model \cite{SMPL:2015} and $G$ as the model of \cite{Hartman_2023_ICCV,hartman2025basis}. In Table \ref{tab:Ablation} we present the experimental results for other combinations of choices for $F$ and $G$. Notably, the performance of the VariShaPE model for these choices is similar to the one presented previously. Moreover, we reported and discussed the results of the MoGeN model using the SMPL latent space. In Table \ref{tab:MoGeNAblation}, we present the results of the MoGeN model where the latent space is instead that of \cite{Hartman_2023_ICCV,hartman2025basis}.  
\begin{table*}[h]
    \centering
    \scriptsize
    \setlength\tabcolsep{2pt}     
    \renewcommand{\arraystretch}{1.0}
    \begin{tabular}{l|cc|cc|}
    &\multicolumn{2}{c|}{Interpolation}&\multicolumn{2}{c}{Extrapolation}\\
    &Linear&MoGeN&Linear&MoGeN\\ \hline
    Mean Error&0.0061&0.0056&0.0173&0.0061\\
    Varifold&0.0072&0.0066&0.0078&0.0069\\
    \end{tabular}
    \caption{Interpolation and Extrapolation Results for the model of \cite{Hartman_2023_ICCV,hartman2025basis}: We present the mean vertex error and mean varifold error of paths in this latent space as produced by the MoGeN model. The reported results are for a test set of mini-sequences extracted from the DFAUST testing set. \label{tab:MoGeNAblation}}
\end{table*}

\section{Conclusions, Limitations and Future Work}
In summary, we have proposed in this paper new self-supervised models for learning not only latent space representations of human body shape and pose but also the underlying geometry of the latent space itself. The latter is achieved  through the introduction of an additional lifted space in which observed body pose motions get unfolded to straight paths. We have further shown how these two frameworks can be used to perform common tasks such as motion interpolation, extrapolation and transfer, and to estimate generative models in the space of human body pose and shape.

An important aspect of the proposed methods is the use of mesh invariant feature representations as inputs of the neural networks which a priori allows them to be applied to raw unregistered and noisy meshes with limited pre-processing needed beyond usual rigid alignment. This includes meshes obtained from different scanning modalities as the result of Fig. \ref{fig:MT} shows with a surface obtained from neural radiance field reconstruction that the authors directly acquired using the camera of their phone. One shortcoming this results points to, however, is the oversimplification of the body shape for subjects too different from those that can be represented in the latent space. We should note that this is primarily a limitation of the latent space used for body shape representation. As the generic framework of this paper is not tied to a specific choice of latent representation, we hope to address such issues in future iterations of this work.
\bibliographystyle{splncs03_unsrt}

\newpage
\appendix
\section{Qualitative Comparison to \cite{huang2021arapreg} for Registered Data.}
\begin{figure}[hb]
    \centering
    \includegraphics[width=.45\textwidth]{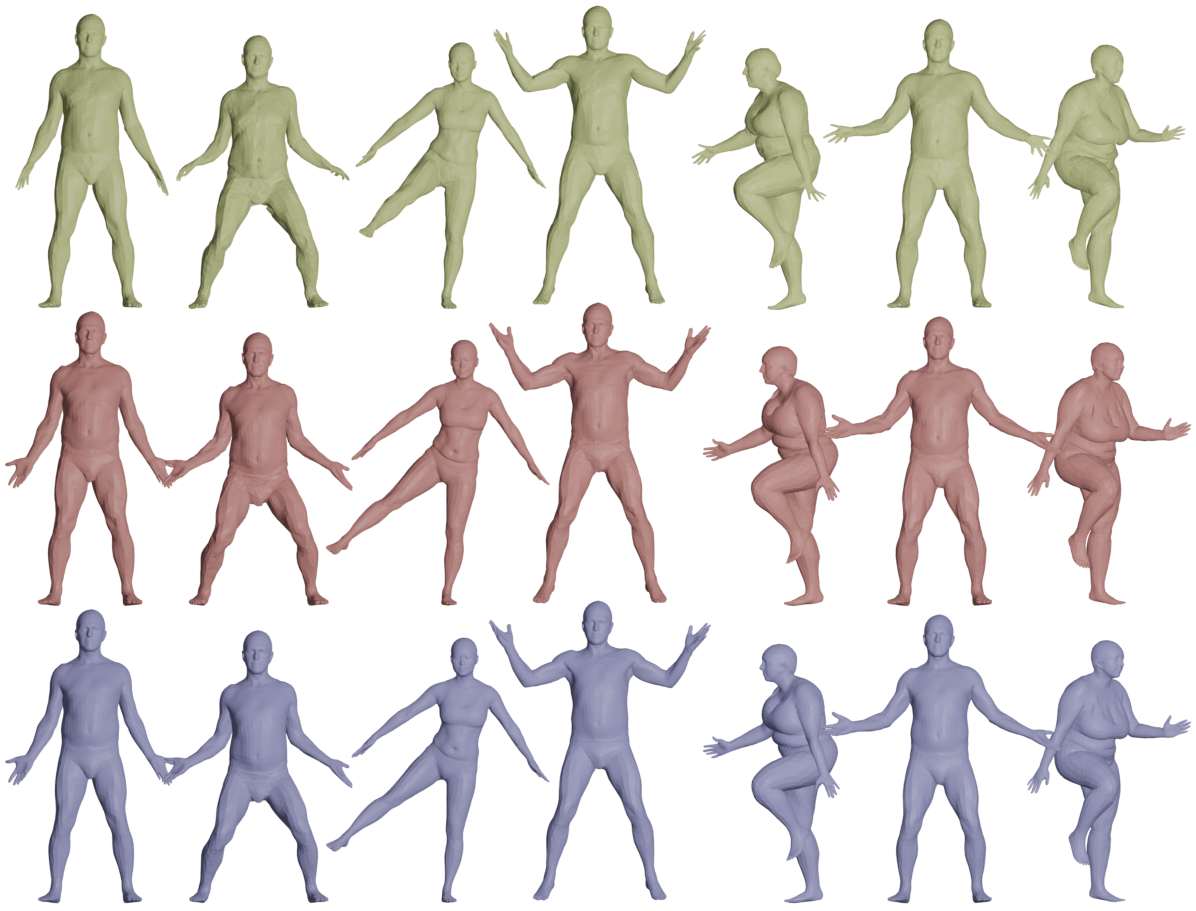}
    \caption{Qualitative Comparison of SMPL latent code retrieval methods for registered data: We display 7 registered human meshes (middle), reconstructions produced by the VAE (top), and reconstructions produced via our model (bottom). \label{fig:ARAP}} 
\end{figure}

Table 1 of the main article reported the quantitative comparison of our model for both registered and unregistered meshes. We further complement those results by emphasizing, in Figure~\ref{fig:ARAP}, the qualitative differences in reconstruction on registered meshes of our proposed method and the variational autoencoder framework described in \cite{huang2021arapreg}.
\section{Qualitative Comparison of MoGeN for Extrapolations.}
\begin{figure}[hb]
    \centering
    \includegraphics[width=.35\textwidth]{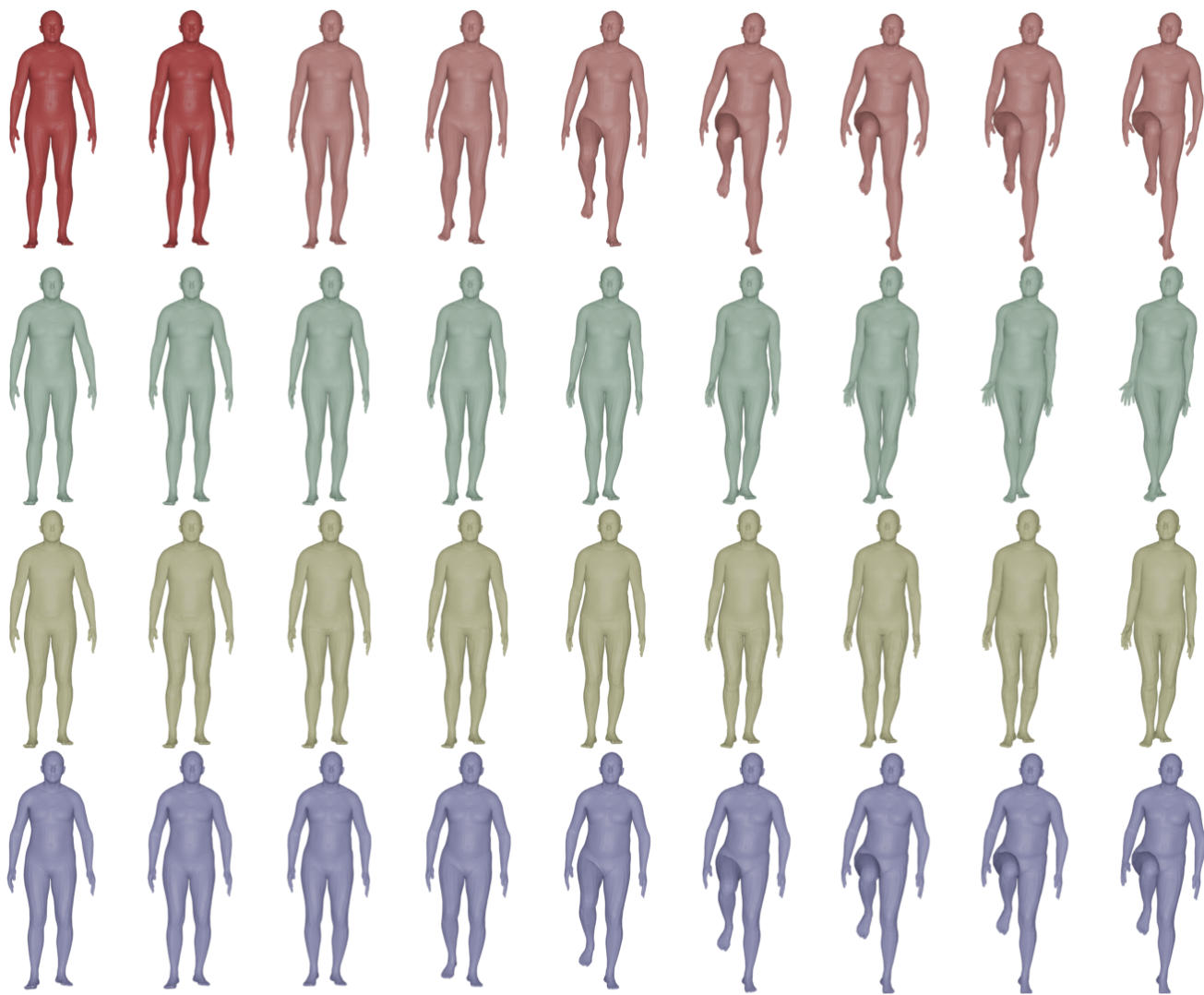}
    \qquad
    \includegraphics[width=.29\textwidth]{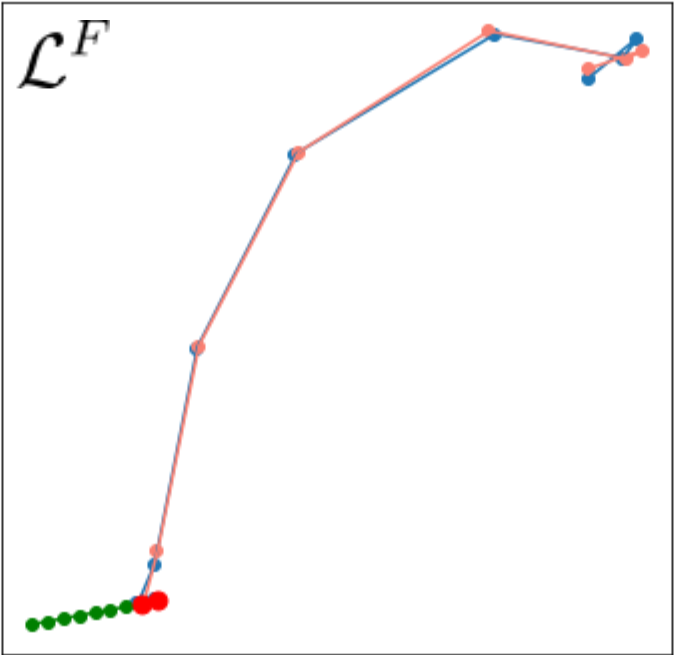}
    \caption{Qualitative comparison of MoGeN with linear extrapolation in SMPL space. In red, we display a sequence of meshes corresponding to a human motion from data with the initial points highlighted. In green, we display the path which arises from linear extrapolation from the initial points in the latent space. Finally, in blue, we display the path produced by the MoGeN extrapolation model with the boundary points as inputs. On the bottom we visualize these paths in two coordinates of the latent space with the same color coding.} \label{fig:extrap}
\end{figure}

In Figure 5 of the main paper, we display the qualitative comparison of MoGeN with linear interpolations in SMPL space. In Figure \ref{fig:extrap} we display a similar comparison of MoGeN with linear extrapolations in SMPL space. This example highlights the strength of our model for learning complex human motions.

\end{document}